\documentclass[runningheads]{llncs}
\usepackage{graphicx}
\usepackage{amsmath,amssymb} 
\usepackage{booktabs}
\usepackage{color}
\usepackage{marvosym}
\usepackage{hyperref}
\usepackage{adjustbox}
\usepackage{cite}
\usepackage[width=122mm,left=12mm,paperwidth=146mm,height=193mm,top=12mm,paperheight=217mm]{geometry}

\begin{document}

\newcommand{\etal}{\textit{et al}.}
\newcommand{\ie}{\textit{i}.\textit{e}., }
\newcommand{\eg}{\textit{e}.\textit{g}.}

\pagestyle{headings}
\mainmatter

\def\fixme#1{\textcolor{red}{[#1]}\marginpar{\textcolor{red}{FIXME}}}
\definecolor{darkgreen}{rgb}{0,0.5,0}
\def\checked#1{#1}

\title{ContextLocNet: Context-Aware Deep Network Models for Weakly Supervised Localization} 

\titlerunning{ContextLocNet}

\authorrunning{V. Kantorov \etal}

\author{Vadim Kantorov, Maxime Oquab, Minsu Cho, and Ivan Laptev}

\institute{WILLOW project team, Inria / ENS / CNRS, Paris, France \\
	\email{\{vadim.kantorov,maxime.oquab,minsu.cho,ivan.laptev\}@inria.fr}
}

\maketitle

\begin{abstract}

We aim to localize objects in images using image-level supervision only.
Previous approaches to this problem mainly focus on discriminative object
regions and often fail to locate precise object boundaries. 
We address this problem by introducing two types of context-aware guidance models, {\em additive
} and {\em contrastive} models, that leverage their surrounding context regions to improve localization. 
The additive model encourages the predicted object region to be
supported by its surrounding context region. 
The contrastive model encourages the
predicted object region to be outstanding from its surrounding context region.
Our approach benefits from the recent success of convolutional neural networks
for object recognition and extends Fast R-CNN to weakly supervised object localization.
Extensive experimental evaluation on the PASCAL VOC 2007 and 2012 benchmarks shows that our context-aware approach significantly improves weakly supervised localization and detection.


\end{abstract}

\section{Introduction}

Weakly supervised object localization and learning
(WSL)~\cite{Wang:2014tg,Cinbis:2015wn} is the problem of localizing spatial
extents of target objects and learning their representations from a dataset with
only image-level labels. 
WSL is motivated by two fundamental issues of conventional object recognition.
First, the strong supervision in terms of object bounding boxes or segmentation
masks is difficult to obtain and prevents scaling-up object localization to
thousands of object classes. Second, imprecise and ambiguous manual annotations
can introduce subjective biases to the learning. 
Convolutional neural networks (CNN)~\cite{LeCun:1989bx,Krizhevsky:2012wl}
have recently taken over the state of the art in many computer vision tasks.
CNN-based methods for weakly supervised object localization have been explored
in~\cite{Oquab:2015us,Bilen:2015uo}.
Despite this progress, WSL remains a very challenging problem. The state-of-the-art performance
of WSL on standard benchmarks~\cite{Wang:2014tg,Cinbis:2015wn,Bilen:2015uo} is considerably lower compared to the
strongly supervised counterparts~\cite{Girshick:2016ig,ren15fasterrcnn,Gidaris:2015cx}.

Strongly supervised detection methods often use contextual information from regions around the object or from the whole 
image~\cite{Torralba:2003wk,Rabinovich:2007wy,Felzenszwalb:2009wx,Girshick:2016ig,Gidaris:2015cx, desai09}:
Indeed, visual context often provides useful information about which image regions are likely to
be a target class according to object-background or object-object relations,
e.g., a boat in the sea, a bird in the sky, a person on a horse, a table around
a chair, etc. However, can a similar effect be achieved for object localization
in a weakly supervised setting, where training data does not contain any
supervisory information neither about object locations nor about context regions?

\begin{figure}[t] 
\includegraphics[width=\textwidth, trim={2mm 6.0cm 2mm 3cm},
clip]{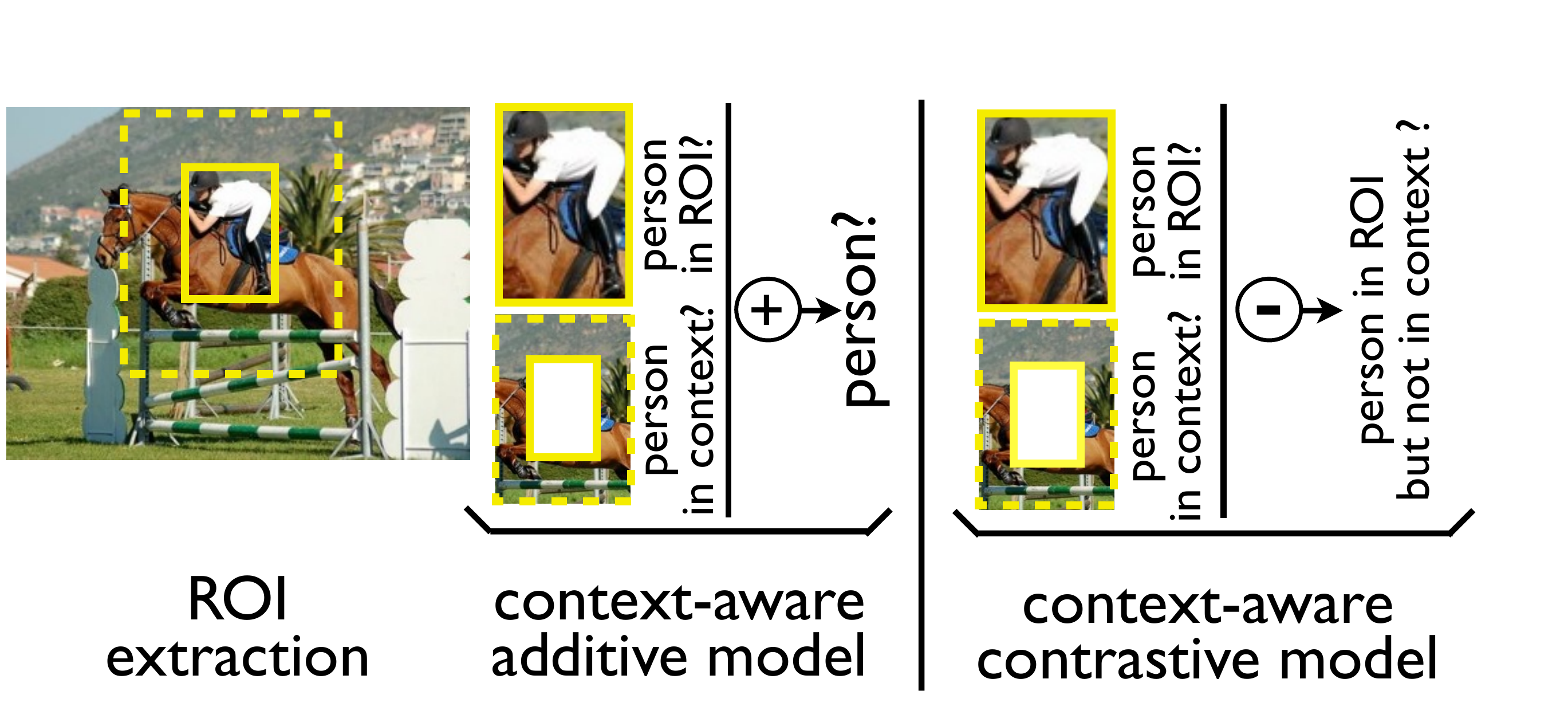} 
\vspace{-6ex} 
\caption[small]{Context-aware guidance for weakly supervised detection.
Given extracted ROIs as localization candidates, our two basic context-aware models, {\em additive} and {\em contrastive} models, leverage their surrounding context regions to improve localization. The additive model relies on semantic consistency that aggregates  class activations from ROI and context. The contrastive model relies on semantic contrast that computes  difference of class activations between ROI and context. For details, see text. (Best viewed in color.)} 
\label{fig:intro} 
\end{figure}

The main contribution of this paper is exploring the use of context as a
supervisory guidance for WSL with CNNs. In a nutshell, we show that, even without strong
supervision, visual context can guide localization in two ways: {\it additive
} and {\it contrastive} guidances.  As the conventional use of contextual
information, the additive guidance enforces the predicted object region to be
compatible with its surrounding context region. This can be encoded by
maximizing the sum of a class score of a candidate region with that of its
surrounding context.
On the other hand, the contrastive guidance encourages the
predicted object region to be outstanding from its surrounding context region.
This can be encoded by maximizing the difference between a class score of the
object region and that of the surrounding context. 
For example, let us consider a candidate box for a person and its surrounding
region of context in Fig.~\ref{fig:intro}. In additive guidance,
appearance of a horse in the surrounding context helps us infer the
surrounded region to contain a person. In contrast guidance, the absence of
target-specific (person) features in its surrounding context helps 
separating the object region from its background.

In this work, we introduce two types of CNN architectures, {\it additive} and
{\it contrastive} models, corresponding to the two contextual guidances. 
Building on the efficient region-of-interest (ROI) pooling
architecture~\cite{ren15fasterrcnn}, the proposed models capture effective
features among potential context regions to localize objects and learn their
representations.
In practice we observe that our additive model prevents expansion of detections
beyond object boundaries. On the other hand, the contrastive model prevents
contraction of detections to small object parts.
In experimental evaluation, we show that our models
significantly outperform the baselines and demonstrate effectiveness of our models for WSL. The project webpage and the code is available at \url{http://www.di.ens.fr/willow/research/contextlocnet}.

\section{Related Work}

In both computer vision and machine learning, there has been a large body of recent research on WSL~\cite{Anonymous:2007fd,shi2012transfer,siva2012defence,Deselaers:2012ci,Siva2013,song2014learning,song14slsvm,bilen2014weakly,bilen2015weakly,Wang:2014tg,Cinbis:2015wn,Oquab:2015us,Bilen:2015uo,Jaderberg:2015vo,Zhou:2015wx}. 
Such methods typically attempt
to localize objects in the form of bounding boxes with visually consistent
appearance in the training images, where multiple objects in different viewpoints
and configurations appear in cluttered backgrounds.  Most of existing approaches
to WSL are formulated as or are closely related to multiple instance learning
(MIL)~\cite{long1998pac}, where each positive image has at least one true
bounding box for a target class, and negative images contain false boxes only.
They typically alternate between
estimating a discriminative representation of the object and selecting an object
box in positive images based on this representation. Since the task consists
in a non-convex optimization problem, WSL has focused on robust initialization
and effective regularization strategies. 

Chum and Zisserman~\cite{Anonymous:2007fd} initialize candidate boxes 
using discriminative visual words, and  update localization by maximizing
the average pairwise similarity across the positive images.
Shi~\etal\cite{shi2012transfer} introduce  
the Latent Dirichlet Allocation (LDA) topic model for WSL, and 
Siva~\etal\cite{siva2012defence} 
propose an effective negative mining approach combined
with discriminative saliency measures.  Deselaers~\etal\cite{Deselaers:2012ci} instead
initialize candidate boxes using the objectness method~\cite{Anonymous:2012kg}, 
and 
propose a CRF-based model that jointly localizes objects in positive training
images. 
Song~\etal formulate 
an initialization strategy for WSL as a discriminative submodular cover problem in
a graph-based framework~\cite{song2014learning}, and develop a negative mining technique to increase robustness against incorrectly localized boxes~\cite{song14slsvm}.  Bilen~\etal
~\cite{bilen2014weakly} propose a relaxed version of MIL that softly
labels object instances instead of choosing the highest scoring ones.
In~\cite{bilen2015weakly}, they also propose a discriminative convex clustering
algorithm to jointly learn a discriminative object model and enforce the
similarity of the localized object regions.
Wang~\etal\cite{Wang:2014tg} propose an iterative latent semantic
clustering algorithm based on latent Semantic Analysis (pLSA) that 
selects the most discriminative cluster for each class in terms of its
classification performance. 
Cinbis~\etal\cite{Cinbis:2015wn} extend a standard MIL approach and propose a
multi-fold strategy that splits the training data to escape bad local optima. 

As CNNs have turned out to be surprisingly effective in many vision tasks
including classification and detection,  recent state-of-the-art WSL approaches
also build on CNN
architectures~\cite{Oquab:2015us,Bilen:2015uo,Jaderberg:2015vo,Zhou:2015wx} or
CNN features~\cite{Wang:2014tg,Cinbis:2015wn}.
Cinbis~\etal\cite{Cinbis:2015wn} combine multi-fold multiple-instance learning
with CNN features. Wang~\etal\cite{Wang:2014tg} develop a semantic
clustering method on top of pretrained CNN features. While these methods produce
promising results, they are not trained end-to-end.
Oquab~\etal\cite{Oquab:2015us} propose a CNN architecture with global max
pooling on top of its final convolutional layer.  Zhou~\etal\cite{Zhou:2015wx}
apply global average pooling instead to encourage the network to cover the full
extent of the object. Rather than directly providing the full extent of the
object, however, these pooling-based approaches are limited to a position of a
discriminative part or require a separate post-processing step to obtain the final
localization. Jaderberg \etal~\cite{Jaderberg:2015vo} propose a CNN architecture
with spatial transformer layers that automatically transform spatial feature
maps to align objects to a common reference frame. Bilen~\etal
~\cite{Bilen:2015uo} modify a region-based CNN
architecture~\cite{Girshick_2015_ICCV} and propose a CNN with two streams, one
focusing on recognition and the other one on localization, that performs
simultaneously region selection and classification.  Our work is related to
these CNN-based MIL approaches that perform WSL by end-to-end training from
image-level labels.  In contrast to the above methods, however, we focus on a
context-aware CNN architecture that exploits contextual relation between a
candidate region and its surrounding regions. 

While contextual information has been widely employed for object
detection~\cite{Oliva:2007ui,Rabinovich:2007wy,Felzenszwalb:2009wx,Girshick:2016ig,Gidaris:2015cx},
the use of context has received relatively little attention in weakly supervised or
unsupervised localization.  Russakovsky~\etal\cite{Russakovsky:2012wj} and
Cinbis~\etal\cite{Cinbis:2015wn} use a background descriptor computed over
features outside a candidate box, and demonstrate that background modelling can
improve WSL as compared to foreground modelling only. 
Doersch~\etal\cite{Gupta:xu09Pf6c} align contextual regions of an object patch
to gradually discovers a visual object cluster in their method of iterative
region prediction and context alignment. Cho~\etal\cite{Cho:2015vz,kwak2015unsupervised} propose a
contrast-based contextual score for unsupervised object localization, which measures the contrast of
matching scores between a candidate region and its surrounding candidate
regions. Our context-aware CNN models are inspired by these previous approaches. 
We would like to emphasize that while the use of contextual information is not new in itself, we apply it to build a novel CNN architecture for WSL, that is, to
the best of our knowledge, unique to our work. We believe that the simplicity of
our basic models makes them extendable to a variety of weakly supervised
computer vision tasks for more accurate localization and learning.

\section{Context-Aware Weakly Supervised Network}

In this section we describe our context-aware deep network for WSL.  Our network
consists of multiple CNN components, each of which builds on previous
models~\cite{Oquab:2015us,Girshick_2015_ICCV,Bilen:2015uo,Gidaris:2015cx}. We
begin by explaining first its overall architecture, and then detail our guidance
models for WSL.

\subsection{Overview}

\begin{figure}[t] \includegraphics[width=\textwidth, trim={1mm 7.3cm 1mm 4cm},
clip]{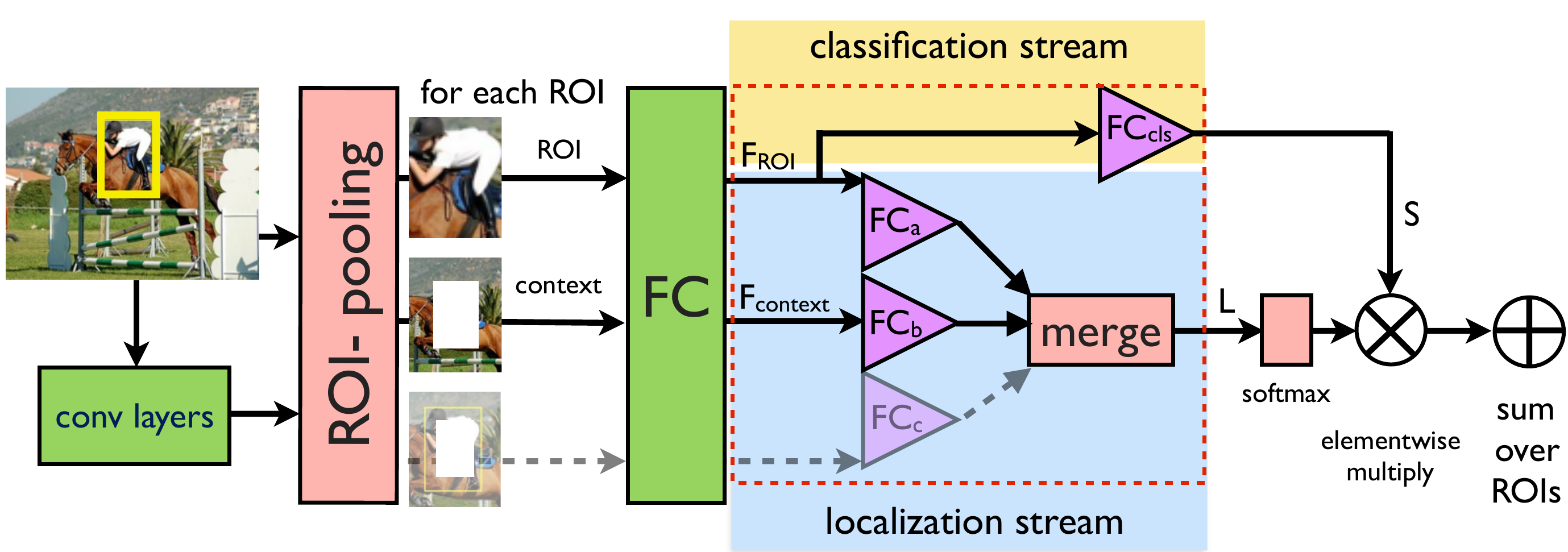} \caption[small]{Our context-aware architecture. 
Convolutional layers and FC layers (in green) correspond to the VGG-F architecture, pre-trained on ImageNet.
The output of FC layers is passed through ReLu to the {\em classification} and {\em localization} streams. 
The classification stream takes features from ROIs, feeds them to a linear layer ${\rm FC_{cls}}$, and outputs classification scores ${\rm S_{ROI}}$. The localization stream takes features from ROIs and their context regions, processes them through our context-aware guidance models, and outputs localization scores ${\rm L_{ROI}}$. The final output is a product of classification and localization scores for each ROI and object class.
${\rm FC_{cls}}$, ${\rm FC_{a}}$, ${\rm FC_{b}}$, ${\rm FC_{c}}$ (in purple) are fully-connected linear layers trained from scratch. See text for more details.
} 
\label{fig:model} 
\end{figure}

Following the intuition of
Oquab~\etal\cite{Oquab:2015us}, our CNN-based approach to WSL learns a network from high-scoring object
candidate regions within a classification training setup. In this approach, the
visual consistency of classes within the dataset allows the network to localize
and learn the underlying objects. The overall network architecture is described
in Fig.~\ref{fig:model}.

\subsubsection{Convolutional and ROI Pooling Layers.} 

Our architecture has 5 convolutional layers, followed by a ROI pooling
layer that extracts a set of feature maps, corresponding to the ROI (object
proposal). The convolutional layers, as our base feature extractor, come from
the VGG-F model~\cite{Chatfield14}. 
 Instead of max pooling typically used to process output of the convolutional layers in conventional
CNNs for classification~\cite{Krizhevsky:2012wl,Oquab:2015us}, however, we
follow the ROI pooling of Fast R-CNN~\cite{Girshick_2015_ICCV}, an efficient
region-based CNN for object detection using object
proposals~\cite{uijlings2013selective}. This network first takes the
entire image as input and applies a sequence of convolutional layers resulting in feature maps (256 feature maps with the effective stride of 16 pixels). The network then contains a ROI-pooling
layer~\cite{He:2014wg}, where ROIs (object proposals) extract corresponding
features from the final convolutional layer. Given a ROI on the image and the
feature maps, the ROI-pooling module projects the ROI on the feature maps, pools
corresponding features with a spatially adaptive grid, and then forwards
them through subsequent fully-connected layers. This architecture allows us to
share computations in convolutional layers for all ROIs in an input image.
Following~\cite{Bilen:2015uo}, in this work, we initialize network layers using the weights of 
ImageNet-pretrained VGG-F model~\cite{Chatfield14}, which is then fine-tuned in training.

\subsubsection{Feature Pooling for Context-Aware Guidance.} 
\begin{figure}[t] \includegraphics[width=\textwidth, trim={1mm 7.5cm 1mm 5cm},
clip]{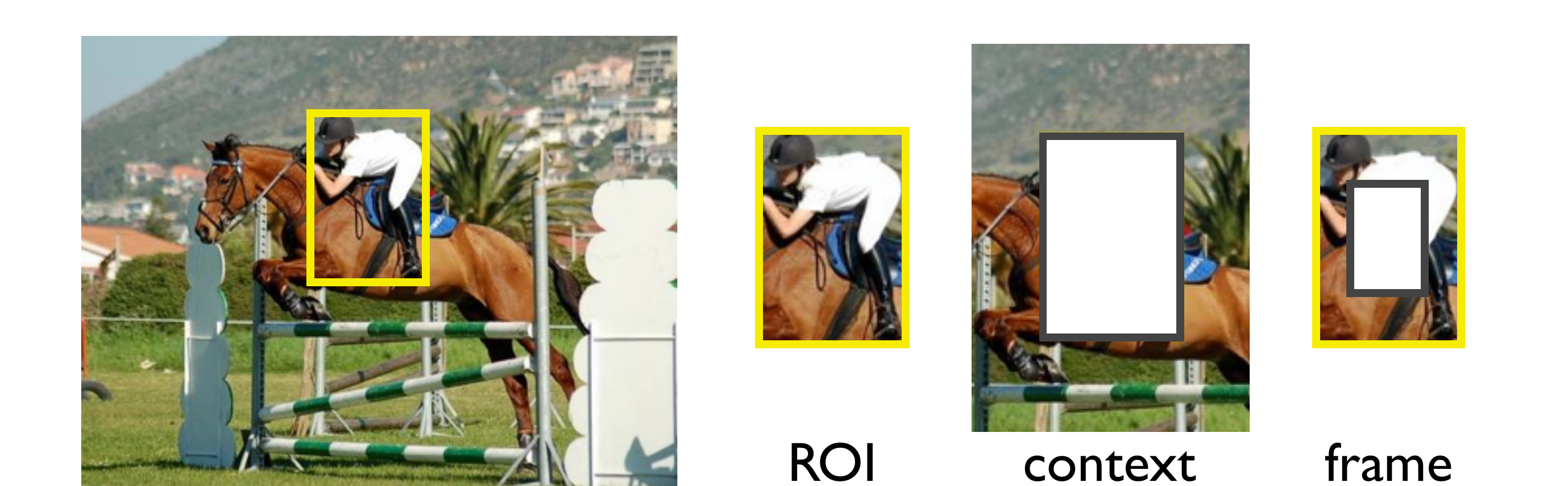} \caption[small]{Region pooling types for our guidance models: ROI pooling, context pooling, and frame pooling. For context and frame, the ratio between the side of the external rectangle and the internal rectangle is fixed as 
$1.8$. Note that context and frame pooling types are designed to produce feature maps of the same shape, \ie frame-shaped feature maps with zeros in the center.} \label{fig:roitransforms} \end{figure}

For context-aware localization and
learning, we extend the ROI pooling by introducing additional pooling types 
for each ROI, in a similar manner to Gidaris~\etal\cite{Gidaris:2015cx}. 
As shown in Fig.~\ref{fig:roitransforms}, 
we define three types of pooling: ROI pooling, context pooling, and frame pooling. 
Given a ROI, \ie ~an object proposal~\cite{uijlings2013selective}, 
the {\it context} is defined as an outer region around the ROI, and the {\it frame} is an inner region ROI. Note that context pooling and frame pooling produce feature maps of the same shape, \ie central area of the outputs will have zero values. As will be explained in Sect.~\ref{sec:contrastive}, this property is useful in our contrast model.
The extracted feature maps are then independently
processed by fully-connected layers (green FC layers in Fig.~\ref{fig:model}), that outputs a ROI feature vector, a context feature vector, and/or a frame feature vector.   
The models will be detailed in Sects.~\ref{sec:additive}
and~\ref{sec:contrastive}. 

\subsubsection{Two-Stream Network.} To combine the guidance model components
with classification, we employ the two-stream architecture of Bilen and
Vedaldi~\cite{Bilen:2015uo}, which branches a localization stream in parallel
with a classification stream, and produces final classification scores by
performing element-wise multiplication between them. In this two-stream strategy, the
classification score of a ROI is reweighted with its corresponding softmaxed
localization score. As illustrated in Fig.~\ref{fig:model}, the {\em classification stream} takes the feature vector  
$F_{\rm ROI}$ as input and feeds it to a linear layer ${\rm FC_{cls}}$, that outputs a set of class
scores $S$. Given $C$ classes, processing $K$ ROIs produces a matrix $S \in \mathbb{R}^{K \times C}$. The {\em localization stream} takes $F_{\rm ROI}$ and $F_{\rm context}$ as inputs, processes them through our guidance models, giving a matrix of localization
scores $L \in \mathbb{R}^{K \times C}$. $L$ is then fed to a softmax layer $ [ \sigma(L) ]_{kc} = \frac{\exp(L_{kc})}{\sum_{k'=1}^{K}{\exp(L_{k'c})}}$ which normalizes the localization scores over the ROIs in the image. 
The final score for each ROI and class is obtained by element-wise multiplication of the corresponding scores $S$ and $\sigma(L)$. 

This procedure is done for each ROI and, as a final step, we sum all the ROI
class scores to obtain the image class scores. During training, we use the hinge
loss function and train the model for multi-label image classification: $$L(w) =
\frac{1}{C \cdot N}\sum_{c=1}^{C}\sum_{i=1}^{N}\max(0, 1 - y_{ci} \cdot f_c(x_i;
w)),$$ where $f_c(x; w)$ is the score of our model evaluated on input image $x$
pararmeterized by $w$ (all weights and biases) for a class $c$; $y_{ci} = 1$ if
$i$'th image contains a ground truth object of class $c$, otherwise $y_{ci} =
-1$. Note that the loss is normalized by the number of classes $C$ and the
number of examples $N$.

\subsection{Additive Model} \label{sec:additive}

\begin{figure}[t] \includegraphics[width=\textwidth, trim={2mm 11.5cm 2mm 3cm},
clip]{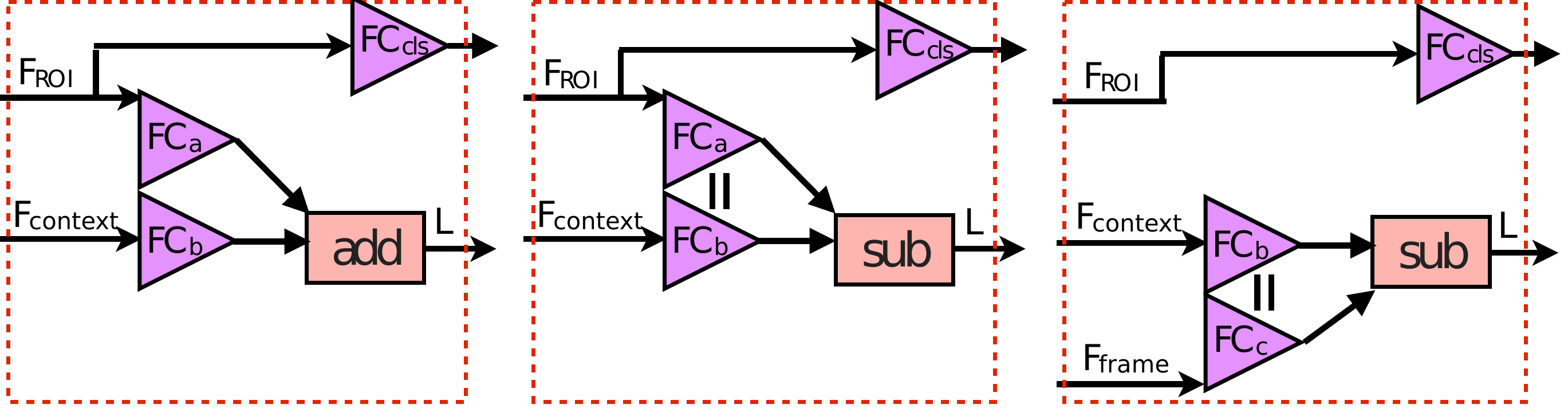} \begin{minipage}{0.32\linewidth} \centering
\small{(a) Additive model. } \end{minipage} \begin{minipage}{0.32\linewidth}
\centering \small{(b) Contrastive model A. } \end{minipage}
\begin{minipage}{0.32\linewidth} \centering \small{(c) Contrastive model S. }
\end{minipage}
\caption[small]{Context-aware guidance models.
The additive model takes outputs of ROI and context pooling, feeds them to independent fully-connected layers, and compute localization scores by adding their outputs.  
The contrastive models take outputs of ROI (or frame) and context pooling, feed them to a shared fully-connected layer (\ie~two fully-connected layers with all parameter shared), and compute localization scores by subtracting the output of context from the other. For details, see the text.}
\label{fig:models} \end{figure}

The additive model, inspired by the conventional use of contextual
information~\cite{Oliva:2007ui,Rabinovich:2007wy,Felzenszwalb:2009wx,Girshick:2016ig,Gidaris:2015cx},
encourages the network to select a ROI that is semantically compatible with its context. 
Specifically, we introduce two fully-connected layers ${\rm FC_{ROI}}$ and ${\rm FC_{context}}$ as shown in Fig.~\ref{fig:models} (a), and the localization score for each ROI is obtained by summing outputs of the layers. Note that compared to context-padding~\cite{Girshick:2016ig}, 
this model separates a ROI and its
context, and learns the adaptation layers ${\rm FC_{ROI}}$ and ${\rm
FC_{context}}$ in different branches. This conjunction of separate
branches allows us to learn context-aware activations for the ROI in an
effective way. 

\begin{figure}[t]
\centering
\includegraphics[width=\textwidth, trim={0mm 3.5cm 8cm 0cm},clip]{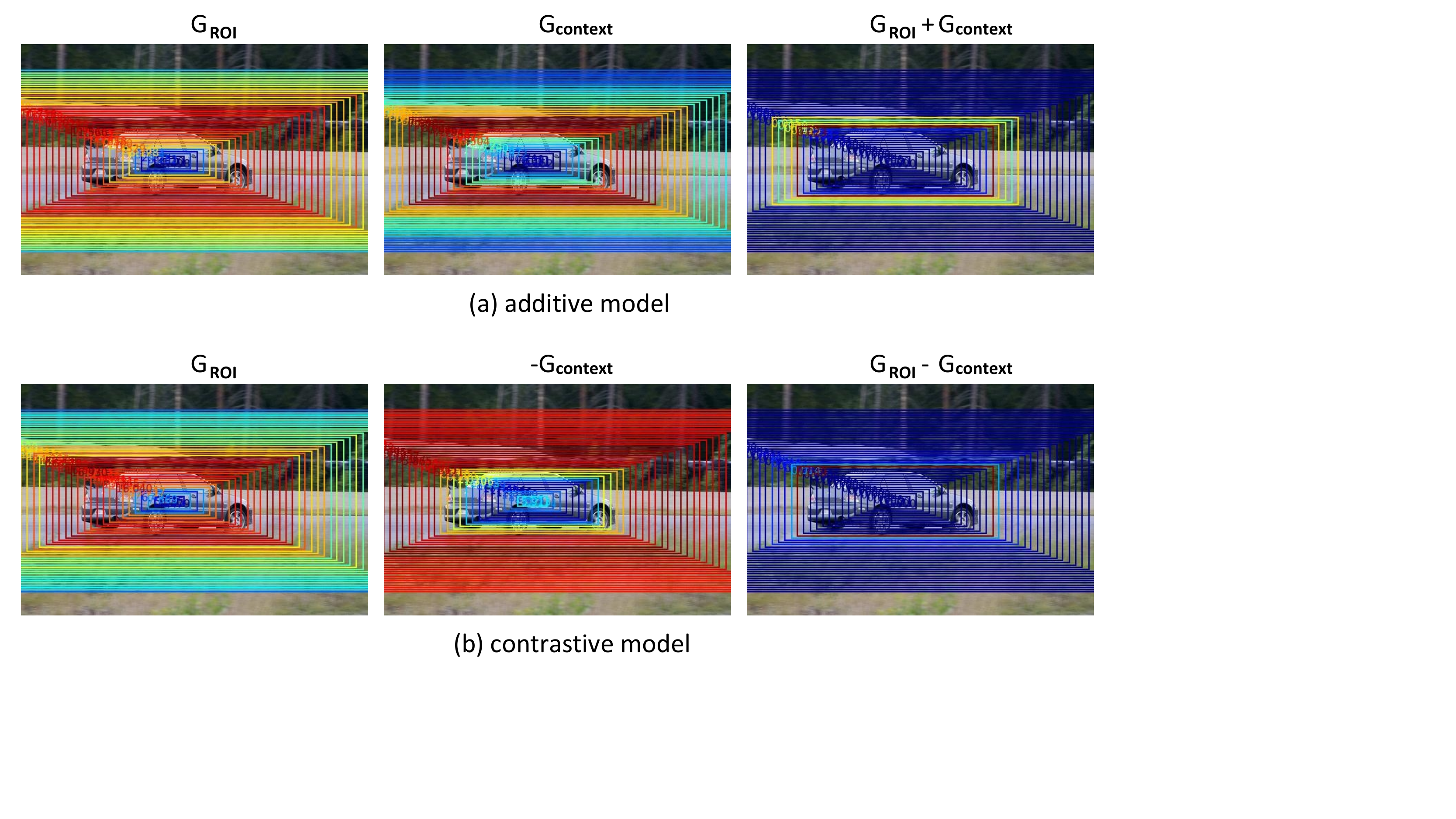}
\caption[small]{
Visualization of object scores produced by different branches of our models. The scores are computed
for the {\em car} class for bounding boxes of different sizes centered on the target object. Red and blue
colors correspond to high and low scores respectively. While the outputs of ${\rm FC_{ROI}}$ branches for the
additive and contrastive models are similar, the ${\rm FC_{context}}$ branches, corresponding to feature
pooling at object boundaries, have notably different behavior. The ${\rm FC_{context}}$ branch of the additive model
discourages detections outside of the object. The ${\rm FC_{context}}$ branch of the contrastive model, discourages
detections on the interior of the object. The combination of the ${\rm FC_{ROI}}$ and ${\rm FC_{context}}$ branches results in correct object
localization for both models.
}
\label{fig:modelvis}
\end{figure}

Figure~\ref{fig:modelvis}(top) illustrates the behavior of the ${\rm FC_{ROI}}$ and ${\rm FC_{context}}$
branches of the additive model trained on PASCAL VOC 2007. The scores of
the target object (car) vary for different sizes of object proposals.
We observe that the ${\rm FC_{context}}$ branch discourages small detections on the
interior of the object as well as large detections outside of object boundaries.
${\rm FC_{context}}$ is, hence, complementary to ${\rm FC_{ROI}}$ and can be expected to prevent detections
outside of objects.

\subsection{Contrastive Model} \label{sec:contrastive}

The contrastive model encourages the network to select a ROI that is
outstanding from its context.  This model is inspired by Cho~\etal's
standout scoring for unsupervised object discovery~\cite{Cho:2015vz}, which
measures the maximum contrast of matching scores between a rectangular box and
its surrounding boxes. We adapt this idea of semantic contrast to our ROI-based
CNN architecture. Specifically, we introduce two fully-connected layers ${\rm FC_{ROI}}$ and ${\rm FC_{context}}$ as shown in Fig.~\ref{fig:models} (b), and the locacalization score for each ROI is obtained by subtracting the output activation of ${\rm FC_{context}}$ from that of ${\rm FC_{ROI}}$ for each ROI.
Note that in order to make subtraction work properly, all weights of the layers ${\rm
FC_{ROI}}$ and ${\rm FC_{context}}$ are shared for this model. Without sharing parameters, this model reduces to the additive model.  

Figure~\ref{fig:modelvis}(bottom) illustrates the behavior of ${\rm FC_{ROI}}$ and ${\rm FC_{context}}$
branches of the contrastive model. We denote by $G_{\rm ROI}$ and $G_{\rm context}$ the outputs of respective layers. The variation of scores for the car object
class and different object proposals indicates low responses of $-G_{\rm context}$
on the interior of the object. The combination $G_{\rm ROI}-G_{\rm context}$ compensate each other
resulting in correct localization of object boundaries. We expect the
contrastive model to prevent incorrect detections on the interior of the object.

One issue in this model is that in the localization stream the shared adaptation layers ${\rm FC_{ROI}}$ and ${\rm FC_{context}}$ need to process input feature maps of different shapes ${\rm F_{ROI}}$ and ${\rm F_{context}}$, \ie  ${\rm FC_{ROI}}$ processes features from a whole region ({\em ROI} in Fig. \ref{fig:roitransforms}), whereas ${\rm FC_{context}}$ processes features from a frame-shaped region ({\em context} in Fig. \ref{fig:roitransforms}). We call this model the asymmetric contrastive model ({\em contrastive A}).  

To remove this asymmetry in the localization stream, we replace ROI pooling with {\em frame} pooling (Fig.~\ref{fig:roitransforms}) that extracts a feature map from an internal rectangular frame of ROI. This allows the shared adaptation layers in the localization stream to process input feature maps of the same shape ${\rm F_{frame}}$ and ${\rm F_{context}}$. We call this model the symmetric contrastive model ({\em contrastive S}). Note that adaptation layer ${\rm
FC_{cls}}$ in the classification stream maintains the original ROI pooling 
regardless of modification in the localization stream. The advantage of this
model will be verified in our experimental section.

\section{Experimental Evaluation}
\subsection{Experimental Setup} 
 \subsubsection{Datasets and Evaluation Measures.}

We evaluate our method on PASCAL VOC 2007 dataset~\cite{Everingham10}, which is
a common benchmark in weakly supervised object detection. This dataset contains
2501 training images, 2510 validation images and 4952 test images, with
bounding box annotations provided for 20 object classes. We use the standard
trainval/test splits. 
We also evaluate our method on PASCAL VOC 2012~\cite{pascal-voc-2012}. VOC 2012 contains the same 
object classes as VOC 2007 and is approximately twice larger in size for both splits.

For evaluation, two performance metrics are used: mAP and CorLoc. 
Detection mAP is evaluated using the standard
intersection-over-union (IoU) criterion defined by \cite{Everingham10}.
Correct localization (CorLoc) \cite{Deselaers:2012ci} is a standard metric 
for measuring localization accuracy on a training set, 
where WSL usually provides one object localization per image for a target class.  
CorLoc is evaluated per-class, only on positive images for
that class, and counts the percentage of images for which the highest-scoring
candidate provided by the method overlaps (IoU $> 0.5$) with a ground truth box.
We evaluate this mAP and CorLoc on the test and trainval splits respectively. 

\subsubsection{Implementation Details.}

ROIs for VOC 2007 are directly provided by the authors of the Selective
Search proposal algorithm~\cite{uijlings2013selective}. For VOC 2012, we use the Selective Search windows computed by Girshick~\etal~\cite{Girshick_2015_ICCV}.
Our implementation is done using Torch~\cite{collobert2011torch7}, and we use the
rectangular frame pooling based on the open-sourced code by Gidaris~\etal~\cite{DBLP:journals/corr/GidarisK15a,Gidaris_2015_ICCV}\footnote{\url{http://github.com/gidariss/locnet}} which 
is itself based on Fast R-CNN~\cite{Girshick_2015_ICCV} code.  
We use the pixel$\rightarrow$features map coordinates transform for 
region proposals from the public implementation of \cite{He:2014wg}\footnote{\url{http://github.com/ShaoqingRen/SPP_net}}, 
with offset parameter set to zero (see the precise procedure in our code online\footnotemark[1]).
All of our models, including our reproduction of WSDDN, use the same transform. 
We use the  ratio between the side of the external rectangle and the internal rectangle fixed to 1.8.\footnote{This choice for the frame parameters follows \cite{DBLP:journals/corr/GidarisK15a,Gidaris_2015_ICCV}, and the ratio is kept same for both context and frame pooling types. We have experimented with different ratios, and observed that results of our method change marginally with increasing the ratio, and drop with decreasing  the ratio.}
Our pretrained network is the VGG-F model~\cite{Chatfield14} ported to Torch using the loadcaffe
package~\cite{zagoruyko2015}.
We train our networks using cuDNN~\cite{chetlur2014cudnn} on an NVidia Titan
X GPU. All layers are fine-tuned. Our training parameters are detailed below.

\subsubsection{Parameters.}

For training, we use stochastic gradient descent (SGD) with momentum 0.9, dampening 0.0 on
examples using a batch size of 1. In our experiments (both training and testing) 
we use all ROIs for an image provided by Selective Search~\cite{uijlings2013selective} that have width and height larger than 20 pixels.
The experiments are run for 30 epochs each. The learning rates are set to
$10^{-5}$ for the first ten epochs, then lowered to $10^{-6}$
until the end of training.
We also use jittering over scales. Images are rescaled randomly into one of the
five following sizes: $800\times 608, 656\times 496, 544\times 400, 960\times
720, 1152\times 864$. Random horizontal flipping is also applied. 

At test time, the scores are evaluated on all scales and flips, then averaged.
Detections are filtered to have a minimum score of $10^{-4}$ and then processed by non-maxima suppression with an overlap threshold of 0.4 prior to mAP calculation.

\subsection{Results and Discussion}
\begin{table*}[t]
\footnotesize
\begin{center}
\begin{tabular}{l@{\hskip 0.5cm}l@{\hskip 0.5cm}cc}
\toprule
& Model &  CorLoc & mAP \\
\midrule
(a)&Cinbis \etal \cite{Cinbis:2015wn} &52.0&30.2\\
(b)&Wang \etal \cite{Wang:2014tg}& 48.5 & 30.9 \\
(c)&Wang \etal + context \cite{Wang:2014tg}&&31.6\\
(d)& WSDDN-SSW-S \cite{Bilen:2015uo}&      & 31.1       \\
(e)& WSDNN-SSW-ENS \cite{Bilen:2015uo} & 54.2 & 33.3\\
\midrule
(f) & WSDDN-SSW-S$^*$& 50.0      & 30.5       \\
(g) &additive & 52.8      & 33.3       \\
(h) &contrastive A & 50.2      & 32.2       \\
(i) &contrastive S & \bf{55.1}      & \bf{36.3}       \\

\bottomrule
\end{tabular}

\vspace{1ex}
\caption{Comparison of our proposed models on PASCAL VOC 2007 with the state of the art, CorLoc (\%) and detection mAP (\%)}
\label{tab:results}
\end{center}
\vspace{-6ex}
\end{table*}

We first evaluate our method on the VOC 2007 benchmark and compare results to the recent methods for weakly-supervised object detecton~\cite{Wang:2014tg,Bilen:2015uo} in Table~\ref{tab:results}.
Specifically, we compare to the WSDDN-SSW-S setup of~\cite{Bilen:2015uo} which, similar to our method, uses VGG-F as a base model and Selective Search Windows object proposals. For fair comparison we also compare results to our re-implementation of WSDDN-SSW-S (row (f) in Table~\ref{tab:results}). The original WSDDN-SSW-S employs an additional softmax in the classification stream and uses binary cross-entropy instead of hinge loss, but we found that these differences to have minor effect on the detection accuracy in our experiments (performance matches up to 1\%, see rows (d) and (f)).

Our best model, contrastive S, reaches 36.3\% mAP and outperforms previous WSL methods using selective search object proposals in rows (a)-(e) of Table~\ref{tab:results}.
Class-specific CorLoc and AP results can be found in Tables~\ref{tab:results_by_class} and \ref{tab:results_by_class_corloc}, respectively.

Bilen~\etal\cite{Bilen:2015uo} experiment with alternative options in terms of EdgeBox object proposals, rescaling ROI pooling activations by EdgeBoxes objectness score, a new regularization term and model ensembling. When combined together, these additions improve result in~\cite{Bilen:2015uo} to 39.3\%. Such improvements are orthogonal to our method and we believe our method will benefit from extensions proposed in~\cite{Bilen:2015uo}. We note that our single contrastive S model (36.3\% mAP) outperforms the ensemble of multiple models using SSW in~\cite{Bilen:2015uo} (33.3\% mAP).

\subsubsection{Context Branch Helps.} 

The additive model (row (g) in Table \ref{tab:results}) improves localization (CorLoc) and detection (mAP)  over those of  the WSDDN-SSW-S$^*$ baseline (row (f)). We also applied a context-padding technique~\cite{Girshick:2016ig} to WSDDN-SSW-S$^*$ by enlarging ROI to include context (in the localization branch). Our additive model (mAP 33.3\%) surpasses the context-padding model (mAP 30.9\%).
Contrastive A also improves localization and detection, but performs slightly worse than the additive model (Table \ref{tab:results}, rows (g) and (h)). 
These results show that processing the context in a separate branch helps 
localization in the weakly supervised setup.

\subsubsection{Contrastive Model with Frame Pooling.}

The basic contrastive model above, contrastive A (see Fig. \ref{fig:models}), processes different shapes of feature maps (${\rm F_{ROI}}$ and ${\rm F_{context}}$) in the localization branch while sharing weights between ${\rm FC_{ROI}}$ and ${\rm FC_{context}}$. To the contrary, contrastive S processes the same shape of feature maps (${\rm F_{frame}}$ and ${\rm F_{context}}$) in the localization branch. As shown in rows (h) and (i) of Table \ref{tab:results}, contrastive~S greatly improves CorLoc and mAP over contrastive~A. Our hypothesis is that, since the weights are shared between the two layers in the the localization
branch, these layers may perform better if they process the same shape of feature maps. Contrastive S obtains such a property by using frame pooling.  
This modification allows us to significantly outperform the baselines (rows (a) - (e) in Table \ref{tab:results}). 
We believe that the model overfits less to the central pixels, achieving better performance.
Per-class results are presented in Tables~\ref{tab:results_by_class}~and~\ref{tab:results_by_class_corloc}.

\begin{table*}[]
\footnotesize
\begin{center}
\begin{adjustbox}{max width=1.02\textwidth}

\begin{tabular}{l@{\hskip 0.5cm}c*{20}cc}
\toprule
Model & aer & bik & brd & boa & btl & bus & car & cat & cha & cow &
tbl & dog & hrs & mbk & prs & plt & shp & sfa & trn & tv & mAP \\
\midrule
Cinbis \etal\cite{Cinbis:2015wn} & 39.3&  43.0&  28.8&  20.4&  8.0&  45.5 & 47.9&  22.1 & 8.4&  33.5&  23.6&  29.2&  38.5&  47.9 & 20.3&  20.0&  35.8&  30.8 & 41.0&  20.1 & 30.2\\
Wang \etal\cite{Wang:2014tg}& 48.8 & 41.0 & 23.6 & 12.1 & 11.1 & 42.7 & 40.9 & 35.5 & 11.1 & 36.6 & 18.4 & 35.3 & 34.8 & 51.3 & 17.2 & 17.4 & 26.8 & 32.8 & 35.1 & 45.6 & 30.9\\
Wang \etal+context\cite{Wang:2014tg} & 48.9&  42.3&  26.1&  11.3&  11.9 & 41.3 & 40.9&  34.7&  10.8&  34.7&  18.8 & 34.4&  35.4&  52.7&  19.1 & 17.4&  35.9 & 33.3&  34.8&  46.5 & 31.6\\
\midrule
WSDDN-SSW-S$^*$	& 49.8 & 50.5 & 30.1 & \bf{12.7} & 11.4 & 54.2 & 49.2 & 20.4 & 1.5 & 31.2 & 27.9 & 18.6 & 32.2 & 49.7 & \bf{22.9} & 15.9 & 25.6 & 27.4 & 38.1 & 41.3 & 30.5 \\
additive		& 	48.7 & 50.7 & 29.5 & 12.3 & \bf{14.1} & \bf{56.5} & 51.7 & 21.1 & \bf{4.0} & 30.0 & 36.5 & 22.5 & 42.6 & 56.2 & 21.5 & \bf{17.5} & \bf{29.5} & 27.0 & 41.3 & \bf{52.3} & 33.3 \\
contrastive A	&	52.8 & 49.6 & 28.9 & 6.8 & 10.9 & 50.4 & 52.2 & \bf{35.0} & 3.2 & 31.4 & 37.6 & 39.7 & 44.1 & 53.4 & 10.7 & 17.4 & 24.2 & 30.9 & 37.8 & 26.9 & 32.2 \\
contrastive S	&	\bf{57.1} & \bf{52.0} & \bf{31.5} & 7.6 & 11.5 & 55.0 & \bf{53.1} & 34.1 & 1.7 & \bf{33.1} & \bf{49.2} & \bf{42.0} & \bf{47.3} & \bf{56.6} & 15.3 & 12.8 & 24.8 & \bf{48.9} & \bf{44.4} & 47.8 & \bf{36.3} \\

\bottomrule
\end{tabular}

\end{adjustbox}
\vspace{1ex}
\caption{Per-class comparison of our proposed models on VOC 2007 with the state of the art, detection AP (\%)}
\label{tab:results_by_class}

\end{center}
\vspace{-6ex}
\end{table*}

\begin{table*}[]
\footnotesize
\begin{center}
\begin{adjustbox}{max width=1.02\textwidth}

\begin{tabular}{l@{\hskip 0.5cm}c*{20}cc}
\toprule
Model &  aer & bik & brd & boa & btl & bus & car & cat & cha & cow &
tbl & dog & hrs & mbk & prs & plt & shp & sfa & trn & tv & avg \\
\midrule
Conbis \etal\cite{Cinbis:2015wn} & 65.3&  55.0&  52.4 & 48.3&  18.2&  66.4 & 77.8 & 35.6 & 26.5 & 67.0&  46.9&  48.4 & 70.5&  69.1&  35.2&  35.2&  69.6 & 43.4&  64.6&  43.7&  52.0\\
Wang \etal\cite{Wang:2014tg}& 80.1 & 63.9 & 51.5 & 14.9 & 21.0 & 55.7 & 74.2 & 43.5 & 26.2 & 53.4 & 16.3 & 56.7 & 58.3 & 69.5 & 14.1 & 38.3 & 58.8 & 47.2 & 49.1 & 60.9 & 48.5\\
\midrule
WSDDN-SSW-S$^*$	& 80.4 & 62.4 & 53.8 & \bf{28.2} & 26.0 & 68.0 & 72.5 & 45.1 & 9.3 & 64.4 & 38.8 & 35.6 & 51.4 & 77.1 & \bf{37.6} & 38.1 & 66.0 & 31.2 & 61.6 & 53.0 & 50.0 \\
additive 		& 78.8 & 66.7 & 52.9 & 25.0 & \bf{26.3} & 68.0 & 73.6 & 44.8 & \bf{14.9} & 62.3 & 45.2 & 46.3 & 61.6 & 82.3 & 35.3 & 39.6 & \bf{69.1} & 30.9 & 62.0 & \bf{69.5} & 52.8 \\
contrastive A 	& 78.8 & 62.7 & 51.1 & 20.2 & 21.8 & 68.5 & 71.6 & \bf{55.8} & 10.3 & \bf{67.8} & 46.8 & 53.7 & 62.2 & 82.3 & 26.0 & \bf{40.7} & 55.7 & 33.6 & 55.5 & 39.4 & 50.2 \\
contrastive S 	& \bf{83.3} & \bf{68.6} & \bf{54.7} & 23.4 & 18.3 & \bf{73.6} & \bf{74.1} & 54.1 & 8.6 & 65.1 & \bf{47.1} & \bf{59.5} & \bf{67.0} & \bf{83.5} & 35.3 & 39.9 & 67.0 & \bf{49.7} & \bf{63.5} & 65.2 & \bf{55.1} \\

\bottomrule
\end{tabular}

\end{adjustbox}
\caption{Per-class comparison of our proposed models on VOC 2007 with the state of the art, CorLoc (\%)}
\label{tab:results_by_class_corloc}

\end{center}
\vspace{-8ex}
\end{table*}

\subsubsection{PASCAL VOC 2012 Results.}
The per-class localization results for the VOC 2012 benchmark using our contrastive model S are summarized in Table~\ref{tab:results_by_class_mAP_voc2012}(detection AP) and Table~\ref{tab:results_by_class_corloc_voc2012}(CorLoc).
We are not aware of other weakly supervised localization methods reporting results on VOC 2012. 

\subsubsection{Observations.}

We have explored several other options and made the following observations.
Training the additive model and the contrastive model in a joint manner (adding the outputs of individual models to compute the localization score that is further processed by softmax) have not improve results in our experiments.
Following Gidaris \etal\cite{Gidaris_2015_ICCV}, we have tried adding other types of region pooling as input to the localization branch, however, this did not improve our results significantly. It is possible that different types of context pooling other than rectangular region pooling can provide improvements. We also found that sharing the weights or replacing the context pooling with the frame pooling in our additive model degrades the performance.

\subsubsection{Qualitative Results.} 
\begin{figure*}
\begin{center}
\includegraphics[trim = 0cm 0.8cm 2.8cm 0cm, clip, width=0.98\linewidth] {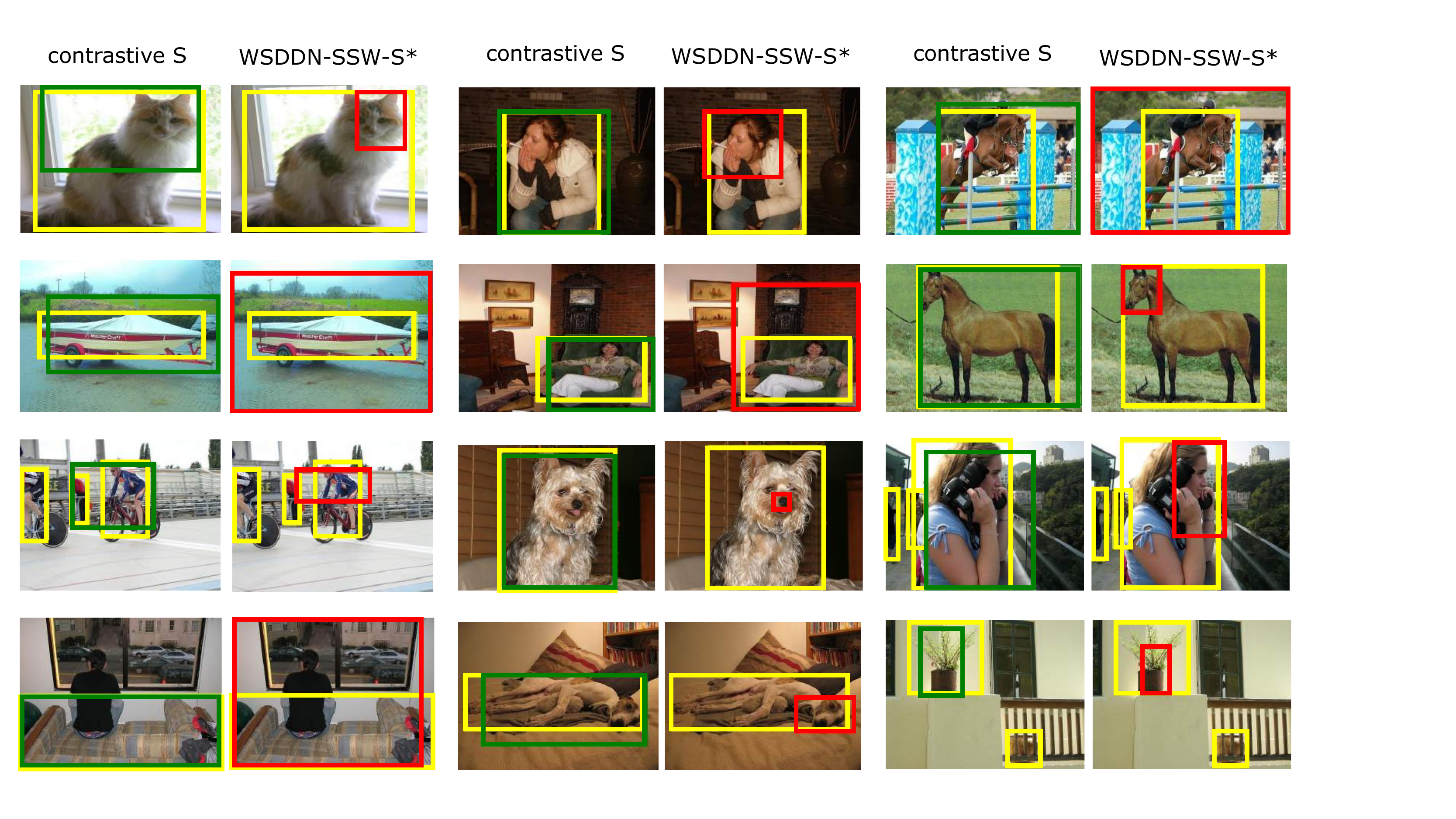}
\includegraphics[trim = 0cm 13.5cm 2.8cm 0cm, clip, width=1.0\linewidth] {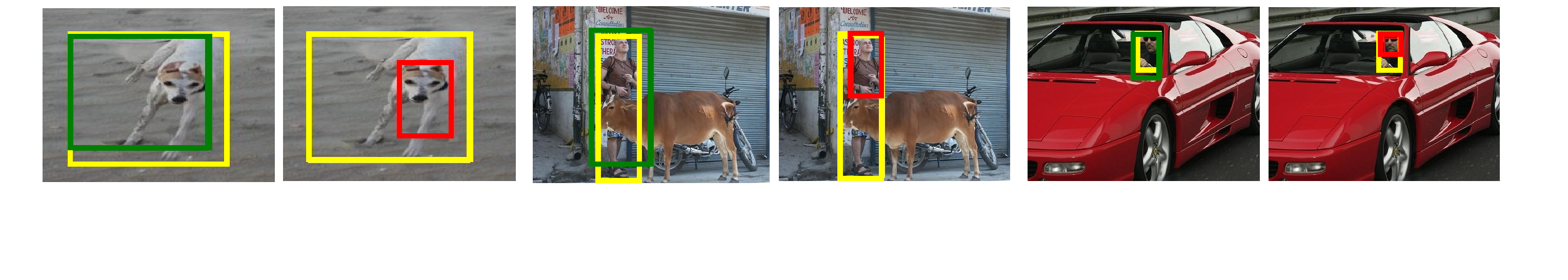}
\includegraphics[trim = 0cm 10cm 3cm 0cm, clip, width=0.985\linewidth] {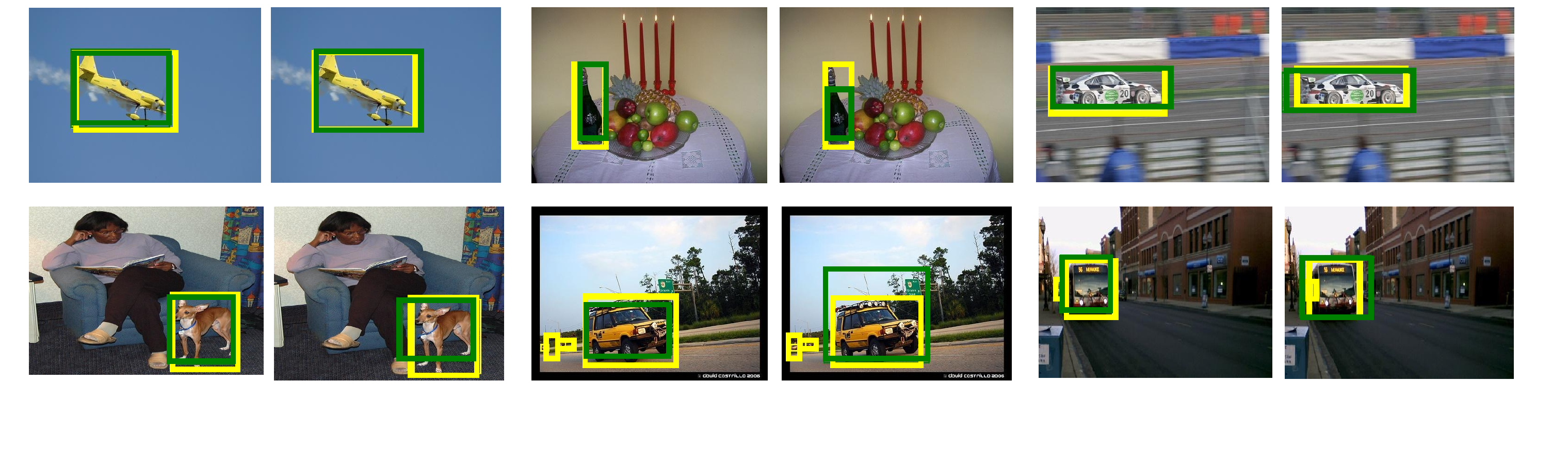}
\includegraphics[trim = 0cm 11cm 3cm 0cm, clip, width=0.98\linewidth] {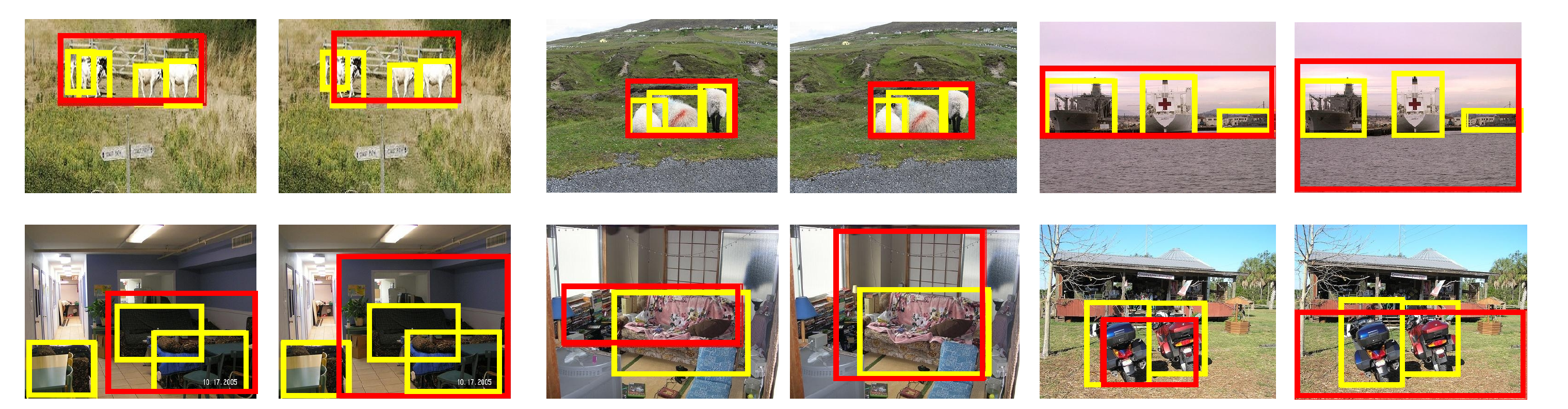}
\mbox{}\vspace{-.3cm}\\
\end{center}
\caption{The first five rows show localization examples where our method (contrastive S) outperforms WSDDN-SSW-S$^*$ baseline. Two next rows show examples where both methods succeed. The last two rows illustrate failure cases for both methods. Our method often suceeds in localizing correct object boundaries on examples where WSDNN-SSW-S$^*$ is locked to descriminative object parts such as heads of people and animals. Typical failure cases for both methods include images with multiple objects of the same class.}
\label{fig:detexamples}
\end{figure*}

We illustrate examples of object detections by our method and WSDDN in Figure~\ref{fig:detexamples}.
We observe that our method tends to provide more accurate localization results for 
classes with localized discriminative parts. For example, for person and animal classes our method often finds
the whole extent of the objects while previous methods tend to localize head regions.
This is consistent with results in Table~\ref{tab:results_by_class} where, for example, the dog class obtains the highest improvement by our contrastive S model when compared to WSDDN. 

Our method still suffers from the second typical failure mode of weakly supervised 
methods, as shown in the two bottom rows of Figure \ref{fig:detexamples}, which is the 
multiple-object case: when many objects of the same class are encountered in close 
vicinity, they tend to be detected as a single object.

\begin{table*}[]
\footnotesize
\begin{center}
\begin{adjustbox}{max width=1.02\textwidth}

\begin{tabular}{l@{\hskip 0.5cm}c*{20}cc}
\toprule
Model &  aer & bik & brd & boa & btl & bus & car & cat & cha & cow &
tbl & dog & hrs & mbk & prs & plt & shp & sfa & trn & tv & mAP \\
\midrule
contrastive S & 64.0 & 54.9 & 36.4 & 8.1 & 12.6 & 53.1 & 40.5 & 28.4 & 6.6 & 35.3 & 34.4 & 49.1 & 42.6 & 62.4 & 19.8 & 15.2 & 27.0 & 33.1 & 33.0 & 50.0 & 35.3 \\

\bottomrule
\end{tabular}

\end{adjustbox}
\vspace{1ex}
\caption{Per-class comparison of the contrastive S model on VOC 2012 test set, AP (\%)}
\label{tab:results_by_class_mAP_voc2012}

\end{center}
\vspace{-.7cm}
\end{table*}

\begin{table*}[]
\footnotesize
\begin{center}
\begin{adjustbox}{max width=1.02\textwidth}

\begin{tabular}{l@{\hskip 0.5cm}c*{20}cc}
\toprule
Model &  aer & bik & brd & boa & btl & bus & car & cat & cha & cow &
tbl & dog & hrs & mbk & prs & plt & shp & sfa & trn & tv & Avg. \\
\midrule

contrastive S & 78.3 & 70.8 & 52.5 & 34.7 & 36.6 & 80.0 & 58.7 & 38.6 & 27.7 & 71.2 & 32.3 & 48.7 & 76.2 & 77.4 & 16.0 & 48.4 & 69.9 & 47.5 & 66.9 & 62.9 & 54.8 \\

\bottomrule
\end{tabular}

\end{adjustbox}
\vspace{1ex}
\caption{Per-class comparison of the contrastive S model on VOC 2012 trainval set, CorLoc (\%)}
\label{tab:results_by_class_corloc_voc2012}

\end{center}
\vspace{-8ex}
\end{table*}

\section{Conclusions}

In this paper, we have presented context-aware deep network models for WSL. 
Building on recent improvements in region-based CNNs, 
we designed a novel localization architecture integrating the idea of contrast-based contextual guidance to the weakly-supervised object localization. We studied the localization component of a weakly-supervised 
detection network and proposed  a subnetwork that effectively makes use of visual contextual information 
that helps 
refining the boundaries of detected objects.
Our results show that the proposed semantic contrast is an effective cue for obtaining more accurate object 
boundaries. Qualitative results show that our method is less sensitive to the typical failure mode of WSL methods, such as 
shrinking to discriminative object parts. Our method has been validated on VOC 2007 and 2012 benchmarks demonstrating
significant improvements over the baselines.

Given the prohibitive cost of large-scale exhaustive annotation, it is crucial to further develop methods for weakly-supervised visual learning. We believe the proposed approach is complementary to many previously explored ideas and could be combined with other techniques to foster further improvements.

\def \imgh {1.5cm}
\def \imgw {1.9cm}

\subsubsection{Acknowledgments.}
We thank Hakan Bilen, Relja Arandjelovi\'{c}, and Soumith Chintala for fruitful discussion and help.
This work was supported by the ERC grants VideoWorld and Activia, and the MSR-INRIA laboratory.

\clearpage

\bibliographystyle{splncs}
\bibliography{shortstrings,contextlocnet_eccv2016}

\begin{thebibliography}{10}

\bibitem{Wang:2014tg}
Wang, C., Ren, W., Huang, K., Tan, T.:
\newblock Weakly supervised object localization with latent category learning.
\newblock In: ECCV.
\newblock Springer (2014)  431--445

\bibitem{Cinbis:2015wn}
Cinbis, R.G., Verbeek, J., Schmid, C.:
\newblock Weakly supervised object localization with multi-fold multiple
  instance learning.
\newblock arXiv preprint arXiv:1503.00949 (2015)

\bibitem{LeCun:1989bx}
LeCun, Y., Boser, B., Denker, J.S., Henderson, D., Howard, R.E., Hubbard, W.,
  Jackel, L.D.:
\newblock Backpropagation applied to handwritten zip code recognition.
\newblock Neural computation \textbf{1}(4) (1989)  541--551

\bibitem{Krizhevsky:2012wl}
Krizhevsky, A., Sutskever, I., Hinton, G.E.:
\newblock Imagenet classification with deep convolutional neural networks.
\newblock In: NIPS. (2012)  1097--1105

\bibitem{Oquab:2015us}
Oquab, M., Bottou, L., Laptev, I., Sivic, J.:
\newblock Is object localization for free?-weakly-supervised learning with
  convolutional neural networks.
\newblock In: CVPR. (2015)  685--694

\bibitem{Bilen:2015uo}
Bilen, H., Vedaldi, A.:
\newblock Weakly supervised deep detection networks.
\newblock In: CVPR. (2016)

\bibitem{Girshick:2016ig}
Girshick, R., Donahue, J., Darrell, T., Malik, J.:
\newblock Region-based convolutional networks for accurate object detection and
  segmentation.
\newblock PAMI \textbf{38}(1) (2016)  142--158

\bibitem{ren15fasterrcnn}
Ren, S., He, K., Girshick, R., Sun, J.:
\newblock Faster r-cnn: Towards real-time object detection with region proposal
  networks.
\newblock In: NIPS. (2015)  91--99

\bibitem{Gidaris:2015cx}
Gidaris, S., Komodakis, N.:
\newblock Object detection via a multi-region and semantic segmentation-aware
  cnn model.
\newblock In: ICCV. (2015)  1134--1142

\bibitem{Torralba:2003wk}
Torralba, A., Murphy, K.P., Freeman, W.T., Rubin, M.A.:
\newblock Context-based vision system for place and object recognition.
\newblock In: ICCV, IEEE (2003)  273--280

\bibitem{Rabinovich:2007wy}
Rabinovich, A., Vedaldi, A., Galleguillos, C., Wiewiora, E., Belongie, S.:
\newblock Objects in context.
\newblock In: ICCV, IEEE (2007)  1--8

\bibitem{Felzenszwalb:2009wx}
Felzenszwalb, P.F., Girshick, R.B., McAllester, D., Ramanan, D.:
\newblock Object detection with discriminatively trained part-based models.
\newblock PAMI \textbf{32}(9) (2010)  1627--1645

\bibitem{desai09}
Desai, C., Ramanan, D., Fowlkes, C.:
\newblock Discriminative models for multi-class object layout.
\newblock In: ICCV. (Sept 2009)  229--236

\bibitem{Anonymous:2007fd}
Chum, O., Zisserman, A.:
\newblock An exemplar model for learning object classes.
\newblock In: CVPR, IEEE (2007)  1--8

\bibitem{shi2012transfer}
Shi, Z., Siva, P., Xiang, T., Mary, Q.:
\newblock Transfer learning by ranking for weakly supervised object annotation.
\newblock In: BMVC. Volume~2., Citeseer (2012) ~5

\bibitem{siva2012defence}
Siva, P., Russell, C., Xiang, T.:
\newblock In defence of negative mining for annotating weakly labelled data.
\newblock In: ECCV.
\newblock Springer (2012)  594--608

\bibitem{Deselaers:2012ci}
Deselaers, T., Alexe, B., Ferrari, V.:
\newblock Weakly supervised localization and learning with generic knowledge.
\newblock IJCV \textbf{100}(3) (2012)  275--293

\bibitem{Siva2013}
Siva, P., Russell, C., Xiang, T., Agapito, L.:
\newblock Looking beyond the image: Unsupervised learning for object saliency
  and detection.
\newblock In: CVPR. (2013)  3238--3245

\bibitem{song2014learning}
Song, H.O., Girshick, R., Jegelka, S., Mairal, J., Harchaoui, Z., Darrell, T.:
\newblock On learning to localize objects with minimal supervision.
\newblock arXiv preprint arXiv:1403.1024 (2014)

\bibitem{song14slsvm}
Song, H.O., Lee, Y.J., Jegelka, S., Darrell, T.:
\newblock Weakly-supervised discovery of visual pattern configurations.
\newblock In: NIPS. (2014)

\bibitem{bilen2014weakly}
Bilen, H., Pedersoli, M., Tuytelaars, T.:
\newblock Weakly supervised object detection with posterior regularization.
\newblock In: BMVC. (2014)

\bibitem{bilen2015weakly}
Bilen, H., Pedersoli, M., Tuytelaars, T.:
\newblock Weakly supervised object detection with convex clustering.
\newblock In: CVPR. (2015)  1081--1089

\bibitem{Jaderberg:2015vo}
Jaderberg, M., Simonyan, K., Zisserman, A.,  et~al.:
\newblock Spatial transformer networks.
\newblock In: NIPS. (2015)  2008--2016

\bibitem{Zhou:2015wx}
Zhou, B., Khosla, A., Lapedriza, A., Oliva, A., Torralba, A.:
\newblock Learning deep features for discriminative localization.
\newblock arXiv preprint arXiv:1512.04150 (2015)

\bibitem{long1998pac}
Long, P.M., Tan, L.:
\newblock Pac learning axis-aligned rectangles with respect to product
  distributions from multiple-instance examples.
\newblock Machine Learning \textbf{30}(1) (1998)  7--21

\bibitem{Anonymous:2012kg}
Alexe, B., Deselaers, T., Ferrari, V.:
\newblock Measuring the objectness of image windows.
\newblock PAMI \textbf{34}(11) (2012)  2189--2202

\bibitem{Girshick_2015_ICCV}
Girshick, R.:
\newblock Fast r-cnn.
\newblock In: ICCV. (2015)  1440--1448

\bibitem{Oliva:2007ui}
Oliva, A., Torralba, A.:
\newblock The role of context in object recognition.
\newblock Trends in cognitive sciences \textbf{11}(12) (2007)  520--527

\bibitem{Russakovsky:2012wj}
Russakovsky, O., Lin, Y., Yu, K., Fei-Fei, L.:
\newblock Object-centric spatial pooling for image classification.
\newblock In: ECCV.
\newblock Springer (2012)  1--15

\bibitem{Gupta:xu09Pf6c}
Doersch, C., Gupta, A., Efros, A.A.:
\newblock Context as supervisory signal: Discovering objects with predictable
  context.
\newblock In: ECCV.
\newblock Springer (2014)  362--377

\bibitem{Cho:2015vz}
Cho, M., Kwak, S., Schmid, C., Ponce, J.:
\newblock Unsupervised object discovery and localization in the wild:
  Part-based matching with bottom-up region proposals.
\newblock In: CVPR. (2015)  1201--1210

\bibitem{kwak2015unsupervised}
Kwak, S., Cho, M., Laptev, I., Ponce, J., Schmid, C.:
\newblock Unsupervised object discovery and tracking in video collections.
\newblock In: ICCV. (2015)  3173--3181

\bibitem{Chatfield14}
Chatfield, K., Simonyan, K., Vedaldi, A., Zisserman, A.:
\newblock Return of the devil in the details: Delving deep into convolutional
  nets.
\newblock In: British Machine Vision Conference. (2014)

\bibitem{uijlings2013selective}
Uijlings, J.R., van~de Sande, K.E., Gevers, T., Smeulders, A.W.:
\newblock Selective search for object recognition.
\newblock IJCV \textbf{104}(2) (2013)  154--171

\bibitem{He:2014wg}
He, K., Zhang, X., Ren, S., Sun, J.:
\newblock Spatial pyramid pooling in deep convolutional networks for visual
  recognition.
\newblock PAMI \textbf{37}(9) (2015)  1904--1916

\bibitem{Everingham10}
Everingham, M., Van~Gool, L., Williams, C.K., Winn, J., Zisserman, A.:
\newblock The pascal visual object classes (voc) challenge.
\newblock IJCV \textbf{88}(2) (2010)  303--338

\bibitem{pascal-voc-2012}
Everingham, M., Van~Gool, L., Williams, C.K.I., Winn, J., Zisserman, A.:
\newblock The {PASCAL} {V}isual {O}bject {C}lasses {C}hallenge 2012 {(VOC2012)}
  {R}esults.
\newblock
  http://www.pascal-network.org/challenges/VOC/voc2012/workshop/index.html

\bibitem{collobert2011torch7}
Collobert, R., Kavukcuoglu, K., Farabet, C.:
\newblock Torch7: A matlab-like environment for machine learning.
\newblock In: BigLearn, NIPS Workshop. Number EPFL-CONF-192376 (2011)

\bibitem{DBLP:journals/corr/GidarisK15a}
Gidaris, S., Komodakis, N.:
\newblock Locnet: Improving localization accuracy for object detection.
\newblock arXiv preprint arXiv:1511.07763 (2015)

\bibitem{Gidaris_2015_ICCV}
Gidaris, S., Komodakis, N.:
\newblock Object detection via a multi-region and semantic segmentation-aware
  cnn model.
\newblock In: ICCV. (2015)  1134--1142

\bibitem{zagoruyko2015}
Zagoruyko, S.:
\newblock loadcaffe.
\newblock \url{https://github.com/szagoruyko/loadcaffe} (2015)

\bibitem{chetlur2014cudnn}
Chetlur, S., Woolley, C., Vandermersch, P., Cohen, J., Tran, J., Catanzaro, B.,
  Shelhamer, E.:
\newblock cudnn: Efficient primitives for deep learning.
\newblock arXiv preprint arXiv:1410.0759 (2014)

\end{thebibliography}
\end{document}